\documentclass[11pt]{article}
\usepackage{coling2020}
\usepackage{times}
\usepackage{url}
\usepackage{latexsym}
\usepackage{enumitem}

\usepackage[pdftex]{graphicx}
\usepackage{multirow}
\usepackage{linguex}

\newcommand{\one}{I}
\newcommand{\two}{I\hspace{-1pt}I}
\newcommand{\three}{I\hspace{-1pt}I\hspace{-1pt}I}
\newcommand{\four}{I\hspace{-1pt}V}
\newcommand{\five}{V}
\newcommand{\six}{V\hspace{-1pt}I}
\newcommand{\seven}{V\hspace{-1pt}I\hspace{-1pt}I}
\newcommand{\eight}{V\hspace{-1pt}I\hspace{-1pt}I\hspace{-1pt}I}

\setlist[itemize]{noitemsep, topsep=0pt}
\setlist[description]{noitemsep, topsep=0pt}

\let\oldexe\exe
\let\oldendexe\endexe
\def\exe{\vspace{-0.5em}\oldexe\setlength{\itemsep}{0pt}} %
\def\endexe{\oldendexe\vspace{-0.5em}}
\let\oldxlist\xlist
\def\xlist{\oldxlist\setlength{\itemsep}{0pt}}

\colingfinalcopy

\title{Modeling and Utilizing User's Internal State\\in Movie Recommendation Dialogue}

\author{Takashi Kodama, Ribeka Tanaka, Sadao Kurohashi \\
    Kyoto University \\
    {\tt \{kodama, tanaka, kuro\}@nlp.ist.i.kyoto-u.ac.jp}}

\date{}

\begin{document}
\maketitle
\begin{abstract}
    Intelligent dialogue systems are expected as a new interface between humans and machines.
    Such an intelligent dialogue system should estimate the user's internal state (UIS) in dialogues and change its response appropriately according to the estimation result.
    In this paper, we model the UIS in dialogues, taking movie recommendation dialogues as examples, and construct a dialogue system that changes its response based on the UIS.
    Based on the dialogue data analysis, we model the UIS as three elements: \textit{knowledge}, \textit{interest}, and \textit{engagement}.
    We train the UIS estimators on a dialogue corpus with the modeled UIS's annotations. The estimators achieved high estimation accuracy.
    We also design response change rules that change the system's responses according to each UIS.
    We confirmed that response changes using the result of the UIS estimators improved the system utterances' naturalness in both dialogue-wise evaluation and utterance-wise evaluation.
\end{abstract}

\section{Introduction}
Dialogue systems using deep learning techniques have attracted much attention, and dialogue research using large-scale data has been developing~\cite{adiwardana2020humanlike,smith-etal-2020-put}.
Currently, mainstream dialogue systems learn to give plausible responses to input utterances superficially.
However, intelligent dialogue systems should understand and interpret the user's internal state (UIS) and utter according to the results of the interpretations.
In this paper, we model the UIS in dialogues and change the system responses based on the UIS.
Based on the analysis of an existing Japanese dialogue data between a movie recommendation dialogue system and humans~\cite{kodama-etal-2019-collection}, we model the UIS in dialogues on the following three axes.
\begin{itemize}
    \item \textbf{Knowledge}: Whether the user is knowledgeable about a topic
    \item \textbf{Interest}: Whether the user is interested in a topic
    \item \textbf{Engagement}: Whether the user is actively engaged in a dialog
\end{itemize}
Understanding the user's knowledge and interest enables the system to provide information and change topics effectively.
The dialogue system can behave appropriately by taking into account the user's engagement. For example, if the user's engagement is high, the system can choose to become a good listener.
These three UISs of knowledge, interest, and engagement are not limited to the domain of movie recommendation and can be applied generically in other dialogues, including chit-chat.

We annotate the movie recommendation dialogue data of 1,000 dialogues with the three modeled UIS by crowdsourcing.
We train the BERT-based UIS estimators on the annotated data.
All the three UIS estimators achieved about 80\% to 85\% accuracy in a 7-point-scale classification when a score gap of $\pm 1$ is allowed.

Moreover, we design and add rules to change the response according to the user's knowledge, interest, and engagement, respectively.
We extend the existing dialogue system to change the response according to the estimated UIS.
The results confirmed that the changes improved the naturalness of the system's utterances in both the dialogue-wise and utterance-wise evaluations.

The contributions of our study are two folds:
\begin{itemize}
    \item We construct a Japanese text dialogue corpus of 1,000 dialogues (10,000 utterances) with the UISs (knowledge, interest, and engagement), which are annotated to user utterances.
    \item We demonstrate the effectiveness of automatic estimation of the UIS and change of system utterances according to the estimation result.
\end{itemize}

\section{Related Work}    \label{sec:related_work}
The UIS in dialogues has been studied from various aspects.

\textit{Emotion} is one of the UISs that has been actively used in dialogue research.
There has been a lot of research on estimating emotions from utterances and generating utterances based on emotions.
\newcite{poria-etal-2019-meld} estimate emotions (e.g., anger, disgust) using verbal and non-verbal information from speeches during the dialogue.
In utterance generation based on emotions, \newcite{zhou-wang-2018-mojitalk} take emoticons in tweets as emotion annotations and proposes a method to generate emotional utterances from a large number of tweets.
\newcite{song-etal-2019-generating} have proposed a method of reflecting specific emotions in utterances, focusing on the explicit and implicit expressions of emotions.

\textit{Persona} focuses on information that cannot be read explicitly from the utterances alone~\cite{li-etal-2016-persona,zhang-etal-2018-personalizing}.
Persona is a speaker's personal background, such as age and gender, as well as the way of speaking based on that background.
Various modeling methods have been proposed to realize consistent personalities, such as IDs~\cite{li-etal-2016-persona}, personal attributes~\cite{ijcai2018-595}, and short personal profile texts~\cite{zhang-etal-2018-personalizing}.
Emotions and personas are mainly used to make responses more informative, but the purpose of our study is to understand the UISs.

\textit{Dialogue acts}~\cite{stolcke-etal-2000-dialogue} are a simple classification of user intentions such as greetings and questions and are often used to understand user's intentions and purposes in task-oriented dialogues.
However, the classification of dialogue acts is highly dependent on the task.
Therefore, there is a problem that the classification needs to be reconsidered when applied to other tasks.
We model UIS in a more general way that can be applied to other task-oriented and chatting dialogues.

The following studies have considered the users' knowledge, interest, and engagement.
\newcite{miyazaki-etal-2013-estimating} investigate effective features for estimating callers' levels of knowledge in call center dialogues and propose a method to estimate their levels of knowledge.

\newcite{Schuller2006RecognitionOI} estimate users' interest level from spoken dialogues.
In text-dialogues, \newcite{inaba-takahashi-2018-estimating} estimate the interest level in 24 topics during chatting dialogues between humans.

The engagement has been actively studied, especially in the field of spoken dialogues.
\newcite{ishihara-2018-willingness} use multimodal information to estimate the user's engagement level from the interview dialogue.
\newcite{Inoue2018} have attempted to estimate engagement using multiple non-verbal behaviors from dialogues between humans and the android robot ERICA~\cite{inoue-etal-2016-talking}.

While these previous studies have addressed the estimation of one of the user's knowledge, interest, and engagement, we deal with these three UISs simultaneously.
Also, we estimate these UISs at each turn of the dialogue, as we expect to change from moment to moment during the dialogues.
We then construct a dialogue system that can interpret the UIS and respond naturally by changing the response appropriately according to the estimation results.

\section{Movie Recommendation Dialogue System with Response Changes}
We extend the existing rule-based movie recommendation dialogue system~\cite{kodama-etal-2019-collection} to change its responses according to UIS.

\subsection{Movie Recommendation Dialogue System}
\subsubsection{Overview}   \label{subsubsec:overview}
\begin{figure}[t!]
    \centering
    \includegraphics[width=\textwidth]{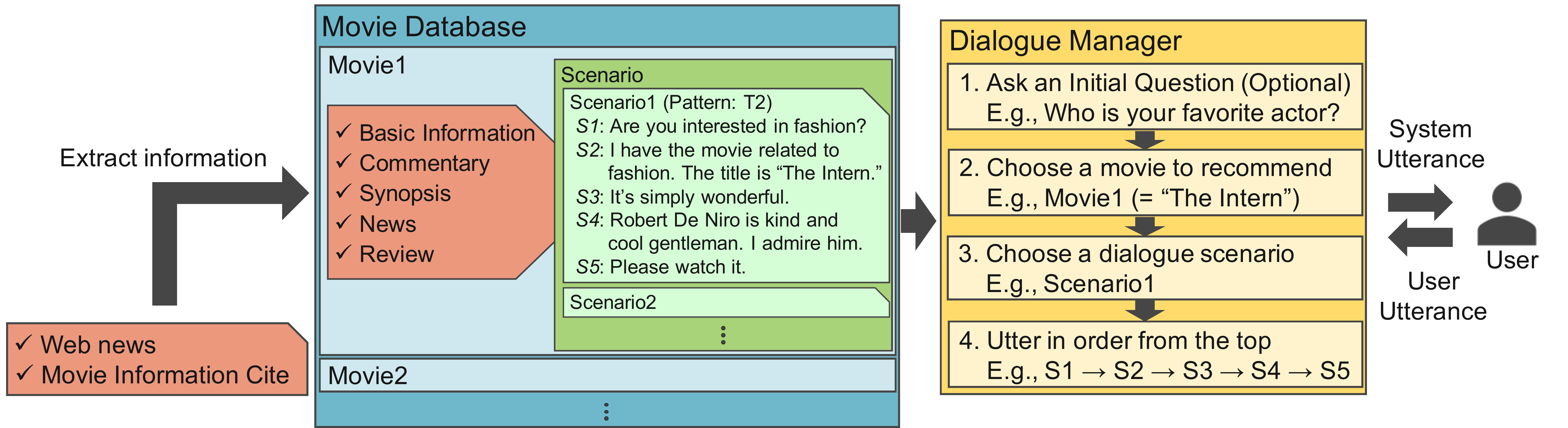}

    \caption{Overview of movie recommendation dialogue system}
    \label{fig:overview}

\end{figure}

Figure \ref{fig:overview} shows an overview of the movie recommendation dialogue system~\cite{kodama-etal-2019-collection}.
The system's utterances are created beforehand for each movie using several templates.
We call the system utterances \textit{scenarios}.
Scenarios are created using information from movie information cites.\footnote{\url{https://movies.yahoo.co.jp/}}
The movie database consists of these scenarios and the movie information for brief question-answering.
The dialogue manager outputs system utterances based on one of the scenarios and also answers simple questions if necessary.
Example~\ref{ex:dialog_example} shows an example of dialogue with this system.\footnote{Note that examples of dialogues presented in this paper are originally in Japanese and were translated by the authors.}
$S$ and $U$ denote system utterances and user utterances, and the numbers next to them indicate the number of turns in the dialogue.
\ex. \label{ex:dialog_example}

    \a.[$S1$:] Are you interested in fashion?
    \b.[$U1$:] I'm interested in it.
    \b.[$S2$:] I have the movie related to fashion. The title is ``The Intern.''
    \b.[$U2$:] I think I've heard the title before.
    \b.[$S3$:] It's simply wonderful.
    \b.[$U3$:] I see.
    \b.[$S4$:] Robert De Niro is kind and cool gentleman. I admire him.
    \b.[$U4$:] It's nice!
    \b.[$S5$:] Please watch it.

There are two methods for choosing a recommended movie: the random method and the initial question method. This ratio is 2:8.
The former method chooses a movie at random.
The latter method asks a user's preference and then chooses a movie according to the answer.
The initial question asking for the preference is randomly selected from the following:
\begin{itemize}
    \item Who is your favorite actor/actress/director?
    \item What is your favorite movie genre?
    \item Which do you like better, Japanese or foreign movies?
\end{itemize}

\subsubsection{Dialogue Scenario}    \label{subsubsec:scenario}
One or more scenarios are created for each movie as shown in Figure \ref{fig:overview}.
Each scenario consists of five utterances ($S1$ to $S5$).
The system starts with a movie-related
topic $S1$.
$S1$ has the following three patterns \textbf{T1}, \textbf{T2} and \textbf{T3}.
The underlined parts depend on movies.
\begin{description}
    \setlength{\leftskip}{1.5em}
    \item[T1: Recent entertainment news] (e.g., It's a hot topic that \underline{the second child was born to actor }\\\underline{Shota Someya and actress Rinko Kikuchi.})
    \item[T2: Movie theme] (e.g., Are you interested in \underline{airplane}?)
    \item[T3: Movie Information (e.g., director's name)] (e.g., Do you know \underline{Hayao Miyazaki}?)
\end{description}
Then, the system presents a recommended movie ($S2$) and successively utters two recommendation points extracted from reviews of each movie ($S3$ and $S4$).
Finally, the system encourages the user to watch the movie ($S5$).
This final utterance is randomly selected from the several prepared utterances (e.g., ``Please watch it.'').

\subsection{UIS-based Response Change} \label{subsec:response_change}
In this section, we describe the rules for changing system responses according to the estimation results of the UIS estimator.
The UIS estimator is created by the following two procedures.
\begin{enumerate}

    \setlength{\itemsep}{0mm}
    \setlength{\parskip}{0mm}
    \setlength{\itemindent}{3mm}
    \item Construct a dialogue corpus by annotating dialogue data~\cite{kodama-etal-2019-collection} with knowledge, interest, and engagement (See Section~\ref{sec:annotation})
    \item Train the UIS estimator using the constructed dialogue corpus (See Section~\ref{subsec:estimation})

\end{enumerate}

The movie recommendation dialogue system follows the scenario, which specifies system utterances in advance.
Therefore, we can guess the target of each UIS (such as ``what users have knowledge about'' and ``what users are not interested in'') from previous system utterances.
We predefine appropriate responses according to the degree of the UIS and its target.
Table \ref{tab:response_change} shows UIS types and the summaries of each response change.
\begin{table}[t!]
    \centering

    \setlength{\tabcolsep}{1mm}
    \scalebox{0.95}{
    \begin{tabular}{l|p{11cm}}
        \hline
        UIS types & Response Change Summaries\\ \hline
        \one. No knowledge of a person & Add brief profile about the person \\
        \two. No knowledge of a movie & Add the movie information (the release year)\\
        \three. Have knowledge of a movie & Change the end of utterance to a consent tone\\
        \four. Have deep knowledge of a movie & Change the utterance to one that assumes the user has watched the movie\\
        \five. No interest in news & Add supplemental description \\
        \six. No interest in a movie theme & Change the utterance to a question that asks the user's preference \\
        \seven. No interest in a person & Change the utterance to a question that asks the user's preference \\
        \eight. No engagement & Change the utterance to a modest tone\\ \hline
    \end{tabular}
    }
    \caption{List of UIS types and Response Change Summaries.}
    \label{tab:response_change}

\end{table}

\subsubsection{Knowledge-based Response Change}  \label{subsubsec:knowledge_response_change}
We prepare the following four types of knowledge-based response change.

\textbf{\one. No knowledge of a person}\quad
If the user doesn't have knowledge immediately after the pattern \textbf{T3} of $S1$, it is considered that the user doesn't know the person (i.e., casts or director).
In this case, we insert a brief profile about the person before $S2$.
We use the first sentence of the Wikipedia article about the person for this profile.
We use MediaWiki API\footnote{\url{https://www.mediawiki.org/wiki/API:Main_page/ja}} to get the Wikipedia article.
Example~\ref{ex:rc_example1} shows an example.
The subscripts $\mathit{before}$ and $\mathit{after}$ denote the case where the response is not changed and the case where the response is changed, respectively.
We underlined the main difference in $S_\mathit{after}$.
The angle bracket in the example shows the UIS that is the basis for judging the response change.
\ex. \label{ex:rc_example1}

    \setlength{\SubExleftmargin}{4em}
    \a.[$S1$:] Do you know George Lucas?
    \b.[$U1$:] I'm not sure.\quad\textbf{$\langle$No knowledge$\rangle$}
    \b.[$S2_\mathit{before}$:] I have the movie directed by George Lucas. The title is ``Star Wars.''
    \b.[$S2_\mathit{after}$:] \underline{George Lucas is an American film director, producer, and screenwriter.} I have ...

\textbf{\two. No knowledge of a movie}\quad
If the user doesn't have knowledge immediately after $S2$, it is assumed that the user doesn't know the recommended movie.
In such a case, we insert the movie information before $S3$, as in Example~\ref{ex:rc_example2}.
We use the released year of the movie as supplemental information.
\ex. \label{ex:rc_example2}

    \setlength{\SubExleftmargin}{4em}
    \a.[$S2$:] The title is ``Honnōji Hotel.''
    \b.[$U2$:] I don't know that movie.\quad\textbf{$\langle$No knowledge$\rangle$}
    \b.[$S3_\mathit{before}$:] The last scene in Kyoto, ...
    \b.[$S3_\mathit{after}$:] \underline{This movie was released in 2017.} The last scene in Kyoto, ...

\textbf{\three. Have knowledge of a movie}\quad
If the user has knowledge immediately after $S2$ or $S3$, the system changes the end of the next utterance ($S3$ or $S4$) to a consent tone (See Example~\ref{ex:rc_example3}).
The response changes in the case of $S3$ and the case of $S4$ are independent.
\ex. \label{ex:rc_example3}

    \setlength{\SubExleftmargin}{4em}
    \a.[$S2$:] The title is ``Erased.''
    \b.[$U2$:] Yes, I watched it on DVD.\quad\textbf{$\langle$Have knowledge$\rangle$}
    \b.[$S3_\mathit{before}$:] This film has a warm message at the base of the story that will impress you!
    \b.[$S3_\mathit{after}$:] This film has a warm message at the base of the story that will impress you, \underline{don't you?}

\textbf{\four. Have deep knowledge of a movie}\quad
As shown in Example~\ref{ex:rc_example4}, if the user has knowledge immediately after all of $S2$, $S3$, and $S4$, we assume that the user has already watched the recommended movie.
The system changes $S5$ to the utterance that assumes the user has watched the movie.
This utterance is randomly selected from the following:
\begin{itemize}
    \item You may want to watch this movie again.
    \item Please watch it again.
\end{itemize}
\ex. \label{ex:rc_example4}

    \setlength{\SubExleftmargin}{4em}
    \a.[$S2$:] The title is ``Cafe Funiculi Funicula.''
    \b.[$U2$:] You know well!\quad\textbf{$\langle$Have knowledge$\rangle$}
    \b.[$S3$:] I think that real actors are great. You feel it again, don't you?
    \b.[$U3$:] Their performances show their age.\quad\textbf{$\langle$Have knowledge$\rangle$}
    \b.[$S4$:] The episode where Kazu-chan and Shintani-kun get closer and succeeds in love is very heartwarming, isn't it?
    \b.[$U4$:] I think so too!\quad\textbf{$\langle$Have knowledge$\rangle$}
    \b.[$S5_\mathit{before}$:] This is an interesting movie and I highly recommend you watch it.
    \b.[$S5_\mathit{after}$:] Please watch it \underline{again}.

\subsubsection{Interest-based Response Change}  \label{subsubsec:interest_response_change}
We prepare the following three types of interest-based response change.

\textbf{\five. No interest in news}\quad
If the user has no interest immediately after the pattern \textbf{T1} of $S1$, it is assumed that the user is not interested in the presented news.
However, it is possible that the user still shows interest in the movie.
Therefore, the system recommends the movie without changing the recommended movie.
In order to reduce the effect of ignoring the user's utterance, the system randomly selects and inserts one of the following utterances before $S2$ (See Example~\ref{ex:rc_example5}).
\begin{itemize}
    \item It seems to be quite well-known.
    \item It seems to be quite a hot topic.
\end{itemize}
\ex. \label{ex:rc_example5}

    \setlength{\SubExleftmargin}{4em}
    \a.[$S1$:] It's a hot topic that Cocomi, the eldest daughter of actor Takuya Kimura and singer Shizuka Kudo, reported that she had been out after a long time on her Instagram.
    \b.[$U1$:] Yes.\quad\textbf{$\langle$No interest$\rangle$}
    \b.[$S2_\mathit{before}$:] Takuya Kimura is starring in the movie ``Blade of the Immortal.''
    \b.[$S2_\mathit{after}$:] \underline{It seems to be quite well-known.} Takuya Kimura is ...

\textbf{\six. No interest in a movie theme}\quad
If the user has no interest immediately after the pattern \textbf{T2} of $S1$, the user is probably not interested in the movie theme.
In such a case, the system changes the recommended movie.
As shown in Example~\ref{ex:rc_example6}, the system asks an initial question to understand the user's preferences.
This initial question is randomly selected from candidates described in Section~\ref{subsubsec:overview}.
\ex. \label{ex:rc_example6}

    \setlength{\SubExleftmargin}{4em}
    \a.[$S1$:] Are you interested in time travel?
    \b.[$U1$:] No, I'm not so interested in it.\quad\textbf{$\langle$No interest$\rangle$}
    \b.[$S2_\mathit{before}$:] I have the movie related to time travel. The title is ``About Time''
    \b.[$S2_\mathit{after}$:] I see. \underline{Then, who is your favorite director?}

\textbf{\seven. No interest in a person}\quad
If the user has no interest immediately after the pattern \textbf{T3} of $S1$, the user is probably not interested in the person.
The system then changes the recommended movie asking an initial question.
In this case, if the system starts the dialogue with the actress (actor/director) name, it asks user's favorite actress (actor/director).
We show an example in Example~\ref{ex:rc_example7}.
\ex. \label{ex:rc_example7}

    \setlength{\SubExleftmargin}{4em}
    \a.[$S1$:] Do you know Sandra Bullock?
    \b.[$U1$:] I know, but I'm not so interested in her.\quad\textbf{$\langle$No interest$\rangle$}
    \b.[$S2_\mathit{before}$:] I have the movie with Sandra Bullock doing a character voice. The title is ``Minions.''
    \b.[$S2_\mathit{after}$:] I see. \underline{Then, who is your favorite actress?}

\subsubsection{Engagement-based Response Change}  \label{subsubsec:engagement_response_change}
We prepare the following one type of engagement-based response change.

\textbf{\eight. No engagement}\quad
If the user doesn't have engagement immediately after $S4$, it is believed that the system's recommendations have been less effective.
Example~\ref{ex:rc_example8} shows an example.
In this case, the system changes $S5$ to a modest tone.
This utterance is randomly chosen from the following.
\begin{itemize}
    \item Trust me. You will like it.
    \item It may be unexpectedly interesting movie.
\end{itemize}

\ex. \label{ex:rc_example8}

    \setlength{\SubExleftmargin}{4em}
    \a.[$S4$:] I can't get over how cute it is to see Pooh's face change from a sad face ...
    \b.[$U4$:] Okay.\quad\textbf{$\langle$No engagement$\rangle$}
    \b.[$S5_\mathit{before}$:] Please watch it.
    \b.[$S5_\mathit{after}$:] \underline{It may be unexpectedly interesting movie.}

\section{UIS Annotated Corpus}    \label{sec:annotation}

\begin{table}[t!]

    \centering
    \begin{tabular}{l|r||l|rr}
    \hline
    & & & System & User \\ \hline
    \# of dialogues & 1,060 & \# of utterances & 6,154 & 5,094\\
    \# of scenarios & 836 & \# of unique utterances & 4,840 & 2,485\\
    Avg \# of turns & 10.6 & \# of morphemes & 163,347 & 20,279\\
    \# of users & 432 & \# of unique morphemes & 5,123 & 1,786\\
    \hline
    \end{tabular}

    \caption{Statistics of dialogue data~\protect\cite{kodama-etal-2019-collection}. The number of users is calculated using worker's IDs.\protect\footnotemark[4] We divide utterances into morphemes using Juman++~\protect\cite{Tolmachev2020a}.}
    \label{tab:dialog_statistics}

\end{table}
\begin{table}[t!]
    \centering
    \setlength{\tabcolsep}{1mm}
    \scalebox{0.95}{
    \begin{tabular}{ccc|p{14.5cm}}
        \hline
        K & I & E & Dialogue \\ \hline
        & & & \textit{S1}: Are you interested in princess?\\ \hline
        2 & 3 & 3 & \textit{U1}: I'm interested in it. \\ \hline
        & & & \textit{S2}: I have the movie related to princess. The title is ``Color Me True'' \\ \hline
        -3 & 3 & 3 & \textit{U2}: What's it about? \\ \hline
        & & & \textit{S3}: I can't tell you the details, but it's a moving movie. I think it's better to enjoy Haruka Ayase's lines and music instead of thoroughly seeking realism. \\ \hline
        0 & 3 & 3 & \textit{U3}: I'm interested in music. \\ \hline
        & & & \textit{S4}: I thought it was not a tragedy but a sad endng. However, it comes to a grand finale with a flip from black and white to full color. \\ \hline
        -2 & 3 & 3 & \textit{U4}: That's a nice development. \\ \hline
        & & & \textit{S5}: Please watch it. \\ \hline
    \end{tabular}
    }
    \caption{Dialogue corpus example. K, I, and E in the left column represent knowledge, interest, and engagement, respectively.}
    \label{tab:dialog_example}

\end{table}
\begin{table}[t!]
    \centering
    \begin{tabular}{c}
        \begin{minipage}[t]{0.53\hsize}
            \centering
            \setlength{\tabcolsep}{1mm}
            \scalebox{0.95}{
            \begin{tabular}{c|rrr}
                \hline
                Score & Knowledge & Interest & Engagement \\ \hline
                 3 & 13.4\% (684) & 20.2\% (1,030) & 20.4\% (1,039) \\
                 2 & 15.3\% (781) & 22.2\% (1,130) & 19.7\% (1,006)\\
                 1 & 15.6\% (793) & 18.2\% (929) & 17.8\% (906)\\
                 0 & 14.4\% (735) & 13.3\% (680) & 14.1\% (716)\\
                 -1 & 15.8\% (807) & 11.4\% (579) & 12.0\% (609)\\
                 -2 & 14.0\% (711) & 8.7\% (443) & 9.5\% (486)\\
                 -3 & 11.4\% (583) & 5.9\% (303) & 6.5\% (332)\\ \hline
            \end{tabular}
            }
            \caption{Distributions of UIS annotations. The numbers in parentheses indicate the number of utterances.}
            \label{tab:statistics_internal_state}
        \end{minipage}
        \hfill
        \begin{minipage}[t]{0.50\hsize}
            \centering
            \begin{tabular}{l|rr}
                \hline
                UIS & \textit{Full} & \textit{Filtered} \\ \hline
                Knowledge & 0.41 & 0.67 \\
                Interest & 0.40 & 0.59 \\
                Engagement & 0.35 & 0.63 \\ \hline
            \end{tabular}
            \caption{Agreements among annotators}
            \label{tab:annotator_agreement}
        \end{minipage}
    \end{tabular}

\end{table}
We construct a UIS annotated corpus based on the Japanese movie recommendation dialogue data~\cite{kodama-etal-2019-collection}.
The statistics of the dialogue data are shown in Table~\ref{tab:dialog_statistics}.
We use crowdsourcing\footnote{\url{http://crowdsourcing.yahoo.co.jp/}} to annotate the UIS to user's utterances.

Three workers annotate each user utterance with knowledge, interest, and engagement on a 3-point scale (1/0/-1, 1 is the best), reading the dialogue contexts.
The sum of the three scores is regarded as the annotation score, which is a 7-point scale from 3 to -3.
Table~\ref{tab:dialog_example} shows a dialogue corpus example and Table~\ref{tab:statistics_internal_state} shows distributions of each UIS.
The scores for interest and engagement tend to be high, but that for knowledge is distributed almost uniformly.

The UIS annotation is probably influenced by the annotators' subjectivity.
Thus, we verify the reliability of the annotations by measuring the agreement among annotators.
We use Krippendorff's $\alpha$~\cite{krippendorff04} in this paper since each utterance is annotated by three workers and the scale type~\cite{Stevens1946} for annotation is considered to be the ordinal scale.
We calculate $\alpha$ using the difference function for the ordinal scale.

The agreements among annotators are shown in column \textit{Full} of Table~\ref{tab:annotator_agreement}.
The values of $\alpha$ are about 0.40 for any UIS.
In general, sociology concludes that reliable data's $\alpha$ exceeds 0.80.
On the other hand, \newcite{Mathieu-native2016} annotate public speaking presentation performances of native and non-native by crowdsourcing and report that $\alpha$ is reasonable at about 0.40 in the subjective rating task of NLP.
Therefore, it is reasonable that our data's $\alpha$, whose data is annotated by crowdsourcing, is about 0.40.

We prepare another data, which is filtered by removing utterances that contain both 1 and -1 annotations.
We call this data \textit{Filtered}.
The amount of data of \textit{Filtered} is about 80\% compared with that of \textit{Full}.
We compare the accuracy of the estimator trained by \textit{Full} and \textit{Filtered}.
For reference, the annotators' agreements of \textit{Filtered} are shown in column \textit{Filtered} of Table~\ref{tab:annotator_agreement}.

\section{Experiment}

We construct a dialogue system that can change its response according to the estimation result of UIS.
First, we train the UIS estimators using the annotated dialogue corpus.
We then incorporate the estimators into the movie recommendation dialogue system so that it can change its response according to the UIS.
Finally, we collect dialogues using the constructed system with response changes and evaluate them.

\begin{figure}[t!]
    \centering
    \includegraphics[width=\textwidth]{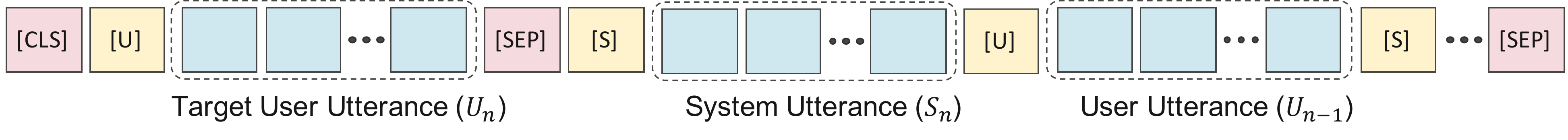}
    \caption{Input format for BERT. [S] and [U] indicate the separation special tokens of system and user utterances, respectively.}
    \label{fig:input}
\end{figure}

\subsection{UIS Estimation}   \label{subsec:estimation}

The UIS estimators are trained for knowledge, interest, and engagement, respectively.

\subsubsection{Estimator}

We use a Japanese pre-trained BERT~\cite{devlin-etal-2019-bert} model\footnote{\url{https://alaginrc.nict.go.jp/nict-bert/index.html}} with BPE~\cite{sennrich-etal-2016-neural} for UIS estimation.
The input format of the target user utterance to BERT is shown in Figure~\ref{fig:input}.
We insert a [CLS] and [SEP] token at the beginning and the end of the target user utterance, respectively.
The dialogue contexts are entered in reverse chronological order.
The separation special tokens [S] and [U] are inserted at the beginning of each system and user utterance.
We use mean squared error as the loss function, and the output is a real value linearly transformed into one dimension from the vector corresponding to the [CLS] token.
\begin{table}[t!]
    \centering
    \scalebox{0.95}{
    \begin{tabular}{l|c|c}
    \hline
    UIS & \textit{Full} & \textit{Filtered} \\ \hline
    Knowledge & 5,094 (4,082/511/501) & 4,073 (3,266/410/397) \\
    Interest & 5,094 (4,082/511/501) & 4,292 (3,424/432/436) \\
    Engagement & 5,094 (4,082/511/501) & 3,926 (3,134/396/396) \\\hline
    \end{tabular}
    }

    \caption{Number of utterances of \textit{Full} and \textit{Filtered}. The numbers in parentheses indicate the number of utterances in training data, development data, and test data, respectively.}

    \label{tab:utterances_data}
\end{table}

\subsubsection{Setting}

We randomly split 1,060 dialogues into 80\%/10\%/10\% for training/development/test.
The number of utterances of each data is shown in Table~\ref{tab:utterances_data}.

We adopt the following two evaluation metrics:
\begin{itemize}
    \item \textbf{Acc}: Percentage of cases where the gap between the estimated score and the correct score is less than or equal to $\pm 0.5$.
    \item \textbf{Broad Acc}: Percentage of cases where the gap between the estimated score and the correct score is less than or equal to $\pm 1.5$.
\end{itemize}
Acc corresponds to the accuracy of 7-point-scale classification, and Broad Acc corresponds to the accuracy of 7-point-scale classification with a score gap of $\pm 1$ allowed.

For fine-tuning, we perform hyperparameter tuning under the following conditions according to the settings of \newcite{devlin-etal-2019-bert}.
We choose the estimator that performs best in Acc on each development set.
The dropout probability was always kept at 0.1.
\begin{itemize}
    \item \textbf{Batch size}: {16, 32}
    \item \textbf{Learning rate (Adam)}: {5e-5, 3e-5, 2e-5}
    \item \textbf{Number of epochs}: {2, 3, 4}
\end{itemize}

\subsubsection{Result}
\begin{table}[t!]
    \centering
    \scalebox{0.95}{
    \begin{tabular}{l|c|c|c|c|c|c}
    \hline
        \multirow{2}{*}{UIS} & \multicolumn{2}{|c|}{$Full_\mathit{Full}$} & \multicolumn{2}{|c|}{$Full_\mathit{Filtered}$} & \multicolumn{2}{|c}{$Filtered_\mathit{Filtered}$} \\ \cline{2-7}
        & Acc(\%) & Broad Acc(\%) & Acc(\%) & Broad Acc(\%) & Acc(\%) & Broad Acc(\%) \\ \hline
        Knowledge & 29.1 & 73.7 & 31.5 & 76.1 & 36.3 & 80.6 \\
        Interest & 32.9 & 82.0 & 36.0 & 85.8 & 33.5 & 83.0 \\
        Engagement & 28.3 & 72.5 & 32.1 & 78.5 & 36.4 & 84.8 \\ \hline
    \end{tabular}
    }

    \caption{Results of UIS estimation}
    \label{tab:internal_state}

\end{table}

Table~\ref{tab:internal_state} shows the estimation results.
$Full_\mathit{Full}$ is an estimator using \textit{Full} for both training and test, and Acc is about 30\% and Broad Acc is about 70-80\% in all UISs.
Considering the majority baselines of knowledge, interest, and engagement are 15.6\%, 22.2\%, and 20.4\%, respectively (See Table \ref{tab:statistics_internal_state}), $Full_\mathit{Full}$ shows reasonable accuracy.
$Full_\mathit{Filtered}$, which is trained on \textit{Full} and is tested on \textit{Filtered}, improves both of Acc and Broad Acc.
This is because the test data for \textit{Filtered} is less noisy and the estimator can estimate UIS more easily.
$Filtered_\mathit{Filtered}$, which is trained and tested on less noisy \textit{Filtered}, achieve the best scores for knowledge and engagement, although this model is a little inferior to $Full_\mathit{Filtered}$ in interest.
The overall accuracy of $Filtered_\mathit{Filtered}$ is about 35\% for Acc and about 80-85\% for Broad Acc.
This result suggests that the data filtering contributes to the improvement of the UIS estimation accuracy.

\subsection{Dialogue-wise and Utterance-wise Evaluation}
\subsubsection{Introduction of UIS Estimation}
We introduce the UIS estimators into the dialogue system.
The system judges whether a user has each of knowledge, interest, and engagement and changes its responses as described in Section \ref{subsec:response_change}.
In experiments, we use $Filtered_\mathit{Filtered}$ estimator placing emphasis on the overall estimation accuracy.
A threshold is set to judge whether a user has knowledge, interest, and engagement.
If it exceeds a positive threshold, the user has its UIS.
If it falls below a negative threshold, the user doesn't have its UIS.
In this paper, the positive and negative thresholds for knowledge and interest are set at $1.5$ and $-1.5$, respectively, and that for engagement are set at $1.0$ and $-1.0$.

\subsubsection{Dialogue-wise Evaluation}
We collected 299 dialogues by crowdsourcing using the movie recommendation dialogue system with response changes (\textit{w-RC}).
We also collected 297 dialogues using the system without response changes (\textit{wo-RC}) for comparison.
In collecting dialogues, we asked workers to answer 5-point Likert-scale questionnaires (5 is the best):
\begin{description}
    \setlength{\itemsep}{-5mm}
    \setlength{\leftskip}{1em}
    \item{(1)} \textbf{PERSUASIVENESS}: The system has made you want to watch the recommended movie.\\
    \item{(2)} \textbf{NATURALNESS}: The system has responded naturally.\\
    \item{(3)} \textbf{SATISFACTION}: The system has satisfied you.\\

\end{description}
Results of the dialogue-wise evaluation by questionnaire are shown in Table~\ref{tab:questionnaire}.
There was no significant difference in PERSUASIVENESS between \textit{w-RC} and \textit{wo-RC}.
On the other hand, \textit{w-RC} was 0.26 points higher than \textit{wo-RC} on NATURALNESS and 0.19 points higher than \textit{wo-RC} on SATISFACTION.
A Wilcoxon rank-sum test was conducted for NATURALNESS and SATISFACTION.
A p value less than 0.05 was considered statistically significant.
The p-values were 0.017 and 0.123, respectively, indicating a significant improvement in NATURALNESS.
The results show that the system's utterances can be made more natural by estimating the UIS and changing the response according to the results.

\begin{table}[t!]
    \centering
    \begin{tabular}{l|r|r}
    \hline
    Question & \textit{w-RC} & \textit{wo-RC} \\ \hline
    PERSUASIVENESS & 3.44 & \textbf{3.48} \\
    NATURALNESS & \textbf{3.46} & 3.20 \\
    SATISFACTION & \textbf{3.34} & 3.15 \\\hline
    \end{tabular}

    \caption{Average scores of dialogue-wise evaluation.}
    \label{tab:questionnaire}

\end{table}

\subsubsection{Utterance-wise Evaluation}
Since the dialogue system used in this paper is scenario-based, the responses of \textit{w-RC} and \textit{wo-RC} are the same, except for the response changes described in Section~\ref{subsec:response_change}.
For each UIS type, we extract a pair of changed and unchanged responses from the collected dialogues and compare them pairwise.
We sample up to 30 pairs for each UIS type and evaluate them by crowdsourcing.
Workers see the dialogue contexts and both of the changed and unchanged responses, and choose the more natural one.
Workers were not told which response was changed, and two additional options were added: ``Both responses are equally natural'' (\textit{Natural}) and ``Both responses are equally unnatural'' (\textit{Unnatural}).
Ten workers evaluated each pair.

\begin{table}[t!]
    \centering
    \scalebox{0.95}{
    \begin{tabular}{l|r|r|r|r}
    \hline
    UIS type & \textit{w-RC} & \textit{wo-RC} & \textit{Natural} & \textit{Unnatural} \\ \hline
    \one. No knowledge of a person (30) & \textbf{240} & 33 & 11 & 16 \\
    \two. No knowledge of a movie (30) & \textbf{180} & 44 & 32 & 44 \\
    \three. Have knowledge of a movie (30) & \textbf{87} & 84 & 50 & 79 \\
    \four. Have deep knowledge of a movie (30) & \textbf{135} & 111 & 11 & 43 \\
    \five. No interest in news (10) & \textbf{54} & 11 & 6 & 29\\
    \six. No interest in a movie theme (30) & \textbf{170} & 71 & 6 & 53 \\
    \seven. No interest in a person (2) & \textbf{11} & 8 & 0 & 1 \\
    \eight. No engagement (30) & \textbf{134} & 100 & 30 & 36\\ \hline
    Overall (192) & \textbf{1011} & 462 & 146 & 301 \\ \hline
    \end{tabular}
    }

    \caption{Result of utterance-wise evaluation for naturalness. The numbers in the table represent the sum of workers' votes. The numbers in parentheses in the UIS type column indicate the number of samples.}
    \label{tab:naturalness_branch_points}

\end{table}

Results of the utterance-wise evaluation are shown in Table~\ref{tab:naturalness_branch_points}.
For all UIS types, the naturalness of the responses by \textit{w-RC} outperformed the responses by \textit{wo-RC}.
This result shows that we have constructed a dialogue system that can provide more natural responses by estimating the proposed UIS and changing the responses according to the estimation results.

\subsubsection{Error Analysis}
For UIS type \three, the response change effect was lower than that of the other types.
This type changes the utterance to a consent tone, but only adds one character to the end of the sentence in Japanese.
We speculate that this small change can not make a large difference in the impression.
Besides, the score of \textit{Unnatural} was as high as \textit{w-RC} and \textit{wo-RC} for this type.
This is probably because the recommendation points extracted from reviews did not fit the contexts.
Example~\ref{ex:rc_error3} is one of the examples, where 6 out of 10 workers judged \textit{Unnatural}.
In this example, the user says, ``I like the earlier movies.'' in $U3$,  but the system ignores that utterance, so the system's utterance seems unnatural.
Future work will improve the method by retrieving recommendation points using dialogue contexts.
\ex. \label{ex:rc_error3}

    \setlength{\SubExleftmargin}{4em}
    \a.[$S3$:] This movie was released in 2015. Directed by Steven Spielberg, written by the Coen brothers and starring Tom Hanks, it's a combination you can't help but expect, and it's a good film that lives up to its expectation as a great movie!
    \b.[$U3$:] I like the earlier movies.
    \b.[$S4_\mathit{before}$:] The Berlin Wall, the bridges over Berlin, the checkpoint, and the streets of East Berlin are all carefully photographed to capture the atmosphere of those days.
    \b.[$S4_\mathit{after}$:] ... all carefully photographed to capture the atmosphere of those days, aren't they?

\section{Conclusion}

In this paper, we modeled the UIS in dialogues in terms of knowledge, interest, and engagement for the appropriate interpretation of the user's utterances.
We also constructed a dialogue corpus by annotating the three modeled UIS into dialogue data by crowdsourcing.
The estimator trained on the dialogue corpus can estimate the UIS with high accuracy based on the user's utterance and dialogue contexts.

Furthermore, we developed a dialogue system that changes the response according to the UIS.
We designed rules to change the response according to the knowledge, interest, and engagement and constructed a rule-based dialogue system to change the responses according to them.
Both results of dialogue-wise and utterance-wise evaluations showed that we constructed a dialogue system that can provide more natural responses using the trained estimator and designed rules.

With the recent development of neural network technology, neural models are required to have \textit{explainability} to clarify their behaviors and the reasons behind the predictions.
We are planning to construct the entire dialogue system with neural network techniques in the future.
We believe that our modeling of the UIS will remain the meaning as a basis for the behavior of the system, and will provide a foothold for research into \textit{explainability}.

\bibliographystyle{coling}
\bibliography{coling2020}

\begin{thebibliography}{}

\bibitem[\protect\citename{Adiwardana \bgroup et al.\egroup
  }2020]{adiwardana2020humanlike}
Daniel De~Freitas Adiwardana, Minh-Thang Luong, David~R. So, Jamie Hall, Noah
  Fiedel, Romal Thoppilan, Zi~Yang, Apoorv Kulshreshtha, Gaurav Nemade, Yifeng
  Lu, and Quoc~V. Le.
\newblock 2020.
\newblock Towards a human-like open-domain chatbot.
\newblock {\em ArXiv}, abs/2001.09977.

\bibitem[\protect\citename{Chollet \bgroup et al.\egroup
  }2016]{Mathieu-native2016}
Mathieu Chollet, Helmut Prendinger, and Stefan Scherer.
\newblock 2016.
\newblock Native vs. non-native language fluency implications on multimodal
  interaction for interpersonal skills training.
\newblock In {\em Proceedings of the 18th ACM International Conference on
  Multimodal Interaction}, pages 386--393.

\bibitem[\protect\citename{Devlin \bgroup et al.\egroup
  }2019]{devlin-etal-2019-bert}
Jacob Devlin, Ming-Wei Chang, Kenton Lee, and Kristina Toutanova.
\newblock 2019.
\newblock {BERT}: Pre-training of deep bidirectional transformers for language
  understanding.
\newblock In {\em Proceedings of the 2019 Conference of the North {A}merican
  Chapter of the Association for Computational Linguistics}, pages 4171--4186.

\bibitem[\protect\citename{Inaba and
  Takahashi}2018]{inaba-takahashi-2018-estimating}
Michimasa Inaba and Kenichi Takahashi.
\newblock 2018.
\newblock Estimating user interest from open-domain dialogue.
\newblock In {\em Proceedings of the 19th Annual {SIG}dial Meeting on Discourse
  and Dialogue}, pages 32--40, Melbourne, Australia.

\bibitem[\protect\citename{Inoue \bgroup et al.\egroup
  }2016]{inoue-etal-2016-talking}
Koji Inoue, Pierrick Milhorat, Divesh Lala, Tianyu Zhao, and Tatsuya Kawahara.
\newblock 2016.
\newblock Talking with {ERICA}, an autonomous android.
\newblock In {\em Proceedings of the 17th Annual Meeting of the Special
  Interest Group on Discourse and Dialogue}, pages 212--215, Los Angeles.

\bibitem[\protect\citename{Inoue \bgroup et al.\egroup }2018]{Inoue2018}
Koji Inoue, Divesh Lala, Katsuya Takanashi, and Tatsuya Kawahara.
\newblock 2018.
\newblock Engagement recognition in spoken dialogue via neural network by
  aggregating different annotators' models.
\newblock In {\em {INTERSPEECH} 2018}, pages 616--620.

\bibitem[\protect\citename{Ishihara \bgroup et al.\egroup
  }2018]{ishihara-2018-willingness}
Takuya Ishihara, Katsumi Nitta, Fuminori Nagasawa, and Shogo Okada.
\newblock 2018.
\newblock Estimating interviewee’s willingness in multimodal human robot
  interview interaction.
\newblock In {\em Proceedings of the 20th International Conference on
  Multimodal Interaction: Adjunct}.

\bibitem[\protect\citename{Kodama \bgroup et al.\egroup
  }2019]{kodama-etal-2019-collection}
Takashi Kodama, Ribeka Tanaka, and Sadao Kurohashi.
\newblock 2019.
\newblock Collection and analysis of meaningful dialogue by constructing a
  movie recommendation dialogue system.
\newblock In {\em Proceedings of the 23rd Workshop on the Semantics and
  Pragmatics of Dialogue - Poster Abstracts}, London, United Kingdom.

\bibitem[\protect\citename{Krippendorff}2004]{krippendorff04}
Klaus Krippendorff.
\newblock 2004.
\newblock {\em Content Analysis: An Introduction to Its Methodology (second
  edition)}.
\newblock Sage Publications.

\bibitem[\protect\citename{Li \bgroup et al.\egroup
  }2016]{li-etal-2016-persona}
Jiwei Li, Michel Galley, Chris Brockett, Georgios Spithourakis, Jianfeng Gao,
  and Bill Dolan.
\newblock 2016.
\newblock A persona-based neural conversation model.
\newblock In {\em Proceedings of the 54th Annual Meeting of the Association for
  Computational Linguistics}, pages 994--1003, Berlin, Germany.

\bibitem[\protect\citename{Miyazaki \bgroup et al.\egroup
  }2013]{miyazaki-etal-2013-estimating}
Chiaki Miyazaki, Ryuichiro Higashinaka, Toshiro Makino, and Yoshihiro Matsuo.
\newblock 2013.
\newblock Estimating callers' levels of knowledge in call center dialogues.
\newblock In {\em {INTERSPEECH} 2013}, pages 2866--2870, Lyon, France.

\bibitem[\protect\citename{Poria \bgroup et al.\egroup
  }2019]{poria-etal-2019-meld}
Soujanya Poria, Devamanyu Hazarika, Navonil Majumder, Gautam Naik, Erik
  Cambria, and Rada Mihalcea.
\newblock 2019.
\newblock {MELD}: A multimodal multi-party dataset for emotion recognition in
  conversations.
\newblock In {\em Proceedings of the 57th Annual Meeting of the Association for
  Computational Linguistics}, pages 527--536, Florence, Italy.

\bibitem[\protect\citename{Qian \bgroup et al.\egroup }2018]{ijcai2018-595}
Qiao Qian, Minlie Huang, Haizhou Zhao, Jingfang Xu, and Xiaoyan Zhu.
\newblock 2018.
\newblock Assigning personality/profile to a chatting machine for coherent
  conversation generation.
\newblock In {\em Proceedings of the Twenty-Seventh International Joint
  Conference on Artificial Intelligence}, pages 4279--4285, Stockholm, Sweden.

\bibitem[\protect\citename{Schuller \bgroup et al.\egroup
  }2006]{Schuller2006RecognitionOI}
Bj{\"o}rn~W. Schuller, Niels K{\"o}hler, Ronald M{\"u}ller, and Gerhard Rigoll.
\newblock 2006.
\newblock Recognition of interest in human conversational speech.
\newblock In {\em {INTERSPEECH} 2006}, pages 793--796.

\bibitem[\protect\citename{Sennrich \bgroup et al.\egroup
  }2016]{sennrich-etal-2016-neural}
Rico Sennrich, Barry Haddow, and Alexandra Birch.
\newblock 2016.
\newblock Neural machine translation of rare words with subword units.
\newblock In {\em Proceedings of the 54th Annual Meeting of the Association for
  Computational Linguistics (Volume 1: Long Papers)}, pages 1715--1725, Berlin,
  Germany.

\bibitem[\protect\citename{Smith \bgroup et al.\egroup
  }2020]{smith-etal-2020-put}
Eric~Michael Smith, Mary Williamson, Kurt Shuster, Jason Weston, and Y-Lan
  Boureau.
\newblock 2020.
\newblock Can you put it all together: Evaluating conversational agents{'}
  ability to blend skills.
\newblock In {\em Proceedings of the 58th Annual Meeting of the Association for
  Computational Linguistics}, pages 2021--2030.

\bibitem[\protect\citename{Song \bgroup et al.\egroup
  }2019]{song-etal-2019-generating}
Zhenqiao Song, Xiaoqing Zheng, Lu~Liu, Mu~Xu, and Xuanjing Huang.
\newblock 2019.
\newblock Generating responses with a specific emotion in dialog.
\newblock In {\em Proceedings of the 57th Annual Meeting of the Association for
  Computational Linguistics}, pages 3685--3695, Florence, Italy.

\bibitem[\protect\citename{Stevens}1946]{Stevens1946}
S.~S. Stevens.
\newblock 1946.
\newblock On the theory of scales of measurement.
\newblock {\em Science}, 103(2684):677--680.

\bibitem[\protect\citename{Stolcke \bgroup et al.\egroup
  }2000]{stolcke-etal-2000-dialogue}
Andreas Stolcke, Klaus Ries, Noah Coccaro, Elizabeth Shriberg, Rebecca Bates,
  Daniel Jurafsky, Paul Taylor, Rachel Martin, Carol Van Ess-Dykema, and Marie
  Meteer.
\newblock 2000.
\newblock Dialogue act modeling for automatic tagging and recognition of
  conversational speech.
\newblock {\em Computational Linguistics}, 26(3):339--374.

\bibitem[\protect\citename{Tolmachev \bgroup et al.\egroup
  }2020]{Tolmachev2020a}
Arseny Tolmachev, Daisuke Kawahara, and Sadao Kurohashi.
\newblock 2020.
\newblock Design and structure of the juman++ morphological analyzer.
\newblock {\em Journal of Natural Language Processing}, 27(1):89--132.

\bibitem[\protect\citename{Zhang \bgroup et al.\egroup
  }2018]{zhang-etal-2018-personalizing}
Saizheng Zhang, Emily Dinan, Jack Urbanek, Arthur Szlam, Douwe Kiela, and Jason
  Weston.
\newblock 2018.
\newblock Personalizing dialogue agents: {I} have a dog, do you have pets too?
\newblock In {\em Proceedings of the 56th Annual Meeting of the Association for
  Computational Linguistics}, pages 2204--2213, Melbourne, Australia.

\bibitem[\protect\citename{Zhou and Wang}2018]{zhou-wang-2018-mojitalk}
Xianda Zhou and William~Yang Wang.
\newblock 2018.
\newblock {M}oji{T}alk: Generating emotional responses at scale.
\newblock In {\em Proceedings of the 56th Annual Meeting of the Association for
  Computational Linguistics}, pages 1128--1137, Melbourne, Australia.

\end{thebibliography}

\end{document}